\documentclass[letterpaper, 10 pt, conference]{ieeeconf}  
\IEEEoverridecommandlockouts                              

\usepackage{graphics} 
\usepackage{epsfig} 
\usepackage{epstopdf}
\usepackage{amsmath} 
\usepackage{amssymb}  
\usepackage{booktabs}
\usepackage{multirow} 
\newcommand{\RNum}[1]{\uppercase\expandafter{\romannumeral #1\relax}}
\usepackage{colortbl}
\usepackage{color, colortbl}
\makeatletter
\let\NAT@parse\undefined
\makeatother
\usepackage{hyperref}
\usepackage{flushend}
\usepackage{threeparttable}
\hypersetup{hidelinks}
\title{\LARGE \bf
  SLAMesh: Real-time LiDAR Simultaneous Localization and Meshing
}

\author{Jianyuan Ruan$^{1}$, Bo Li$^{2}$, Yibo Wang$^{1}$, and Yuxiang Sun$^{1,*}$
\thanks{This work was supported by the HK PolyU Start-up Fund under Grant P0034801, which is awarded to Dr. Yuxiang Sun.}
\thanks{$^{1}$The Hong Kong Polytechnic University, Hung Hom, Hong Kong (e-mail: jianyuan.ruan@connect.polyu.hk; me-yibo.wang@connect.polyu.hk; yx.sun@polyu.edu.hk, sun.yuxiang@outlook.com).}
\thanks{$^{2}$Zhejiang University of Science and Technology, Hangzhou, China (e-mail: 120052@zust.edu.cn).}
\thanks{$^*$Corresponding author: Yuxiang Sun.}}

\begin{document}

\maketitle
\thispagestyle{empty}
\pagestyle{empty}

\begin{abstract}
Most current LiDAR simultaneous localization and mapping (SLAM) systems build maps in point clouds, which are sparse when zoomed in, even though they seem dense to human eyes. Dense maps are essential for robotic applications, such as map-based navigation. Due to the low memory cost, mesh has become an attractive dense model for mapping in recent years. However, existing methods usually produce mesh maps by using an offline post-processing step to generate mesh maps. This two-step pipeline does not allow these methods to use the built mesh maps online and to enable localization and meshing to benefit each other. To solve this problem, we propose the first CPU-only real-time LiDAR SLAM system that can simultaneously build a mesh map and perform localization against the mesh map. A novel and direct meshing strategy with Gaussian process reconstruction realizes the fast building, registration, and updating of mesh maps. We perform experiments on several public datasets. The results show that our SLAM system can run at around $40$Hz. The localization and meshing accuracy also outperforms the state-of-the-art methods, including the TSDF map and Poisson reconstruction. Our code and video demos are available at: \href{https://github.com/lab-sun/SLAMesh}{https://github.com/lab-sun/SLAMesh}.
\end{abstract}

\section{Introduction}

Simultaneous localization and mapping (SLAM) aims to estimate sensor poses and reconstruct traversed environments at the same time. 3-D LiDAR SLAM has attracted significant attention in the robotics research community in recent years. Usually, LiDAR SLAM produces a point-cloud map, which is a basic representation for 3-D environment models \cite{zhang2017low}. With such map type, registration and fusion of new range data can be straightforwardly operated \cite{besl1992method}. Unfortunately, a point-cloud map requires extra data structures, such as KD-tree, to speed up the query of the unordered points. The memory cost also grows significantly with the increased scale or density. Moreover, although point-cloud map seems dense to human eyes, it would still be sparse when zoomed in. Due to the sparsity, it usually requires post-processing operations before the use for navigation, such as voxelization, to produce dense voxel maps. 

\begin{figure}[t]
	\centering
	\includegraphics[scale = 1.1]{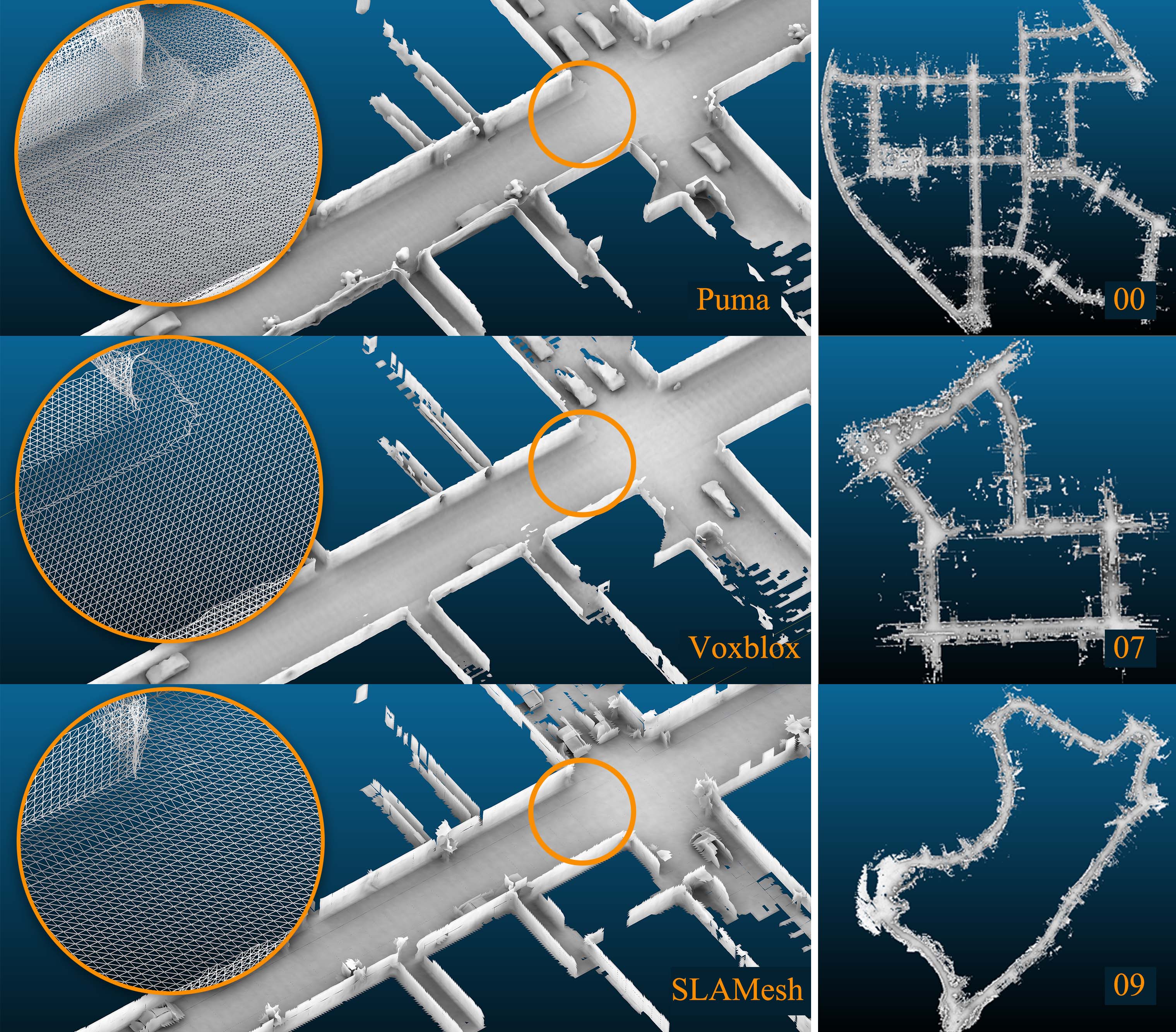}
	\caption{Left: Mesh maps produced by SLAMesh, Puma \cite{vizzo2021poisson} and Voxblox \cite{oleynikova2017voxblox} on the Mai City dataset \cite{vizzo2021poisson}. A zoomed view in each map highlights the connections of the mesh edges. The mesh map of Puma is water-tight but with relatively complex multi-layers. The mesh maps produced by Voxblox and our SLAMesh are of high orderliness, and the vertices are evenly distributed. Right: Mesh maps built by our SLAMesh on the KITTI sequences 00, 07, and 09 demonstrate the accuracy and scalability of SLAMesh in large-scale environments.}
	\label{figmeshsample} 
 \vspace{-0.4cm}

\end{figure}

In this work, we consider a dense map as a kind of representation that models 3-D objects in terms of surfaces or volumes. Under this definition, there exist several types of dense maps. For instance, occupancy grid maps \cite{hornung2013octomap} and truncated signed distance function (TSDF) maps \cite{curless1996volumetric} divide the 3-D space into voxels, making them applicable for robot navigation \cite{oleynikova2017voxblox}. However, the discretization introduces the scalability problem and limits the localization accuracy to the grid size. The normal distribution transform (NDT) \cite{magnusson2007scan}\cite{yokozuka2021litamin2} and surfel maps \cite{behley2018efficient}\cite{chen2019suma++}\cite{Elasticity++} can also be seen as dense maps. They use parameterized representations to describe local structures. Parameterization ensures memory efficiency but limits the ability to describe fine details unless a high resolution is used. Besides, gaps exist around the NDT and surfel map elements. 

Mesh is a 3-D solid surface comprised of faces, edges, and vertices. In the field of 3-D modeling, the triangular mesh has become the dominating representation because it is simple and can approximate most complex 3-D structures. Unlike grid maps, mesh maps alleviate the discretization problem and can model smooth surfaces for robotics applications. Mesh can also depict manifold structures compared to surfel and NDT maps. In addition, mesh map is memory efficient, scalable, and retains the topological information \cite{putz2021mesh}\cite{chen2021range}.

However, real-time 3-D LiDAR mesh mapping is still hard to realize now, especially in large-scale environments. The main difficulty is that mesh building and updating are usually time-consuming. A widely-adopted solution to this problem is first focusing on point-cloud mapping and then performing meshing separately using the point-cloud maps. This two-step solution could not enable localization and meshing to benefit each other \cite{durrant2006simultaneous}.  

To address this issue, this work designs a simultaneous localization and meshing system named SLAMesh. The system can run in real time with only a CPU. In SLAMesh, we make the vertices uniformly distributed, enabling mesh building, registration, and updating more efficiently. Fig. \ref{figmeshsample} shows such a mesh map constructed by the Gaussian process (GP) \cite{rasmussen2003gaussian} reconstruction. Our previous works \cite{li2020gp}\cite{ruan2020gp} have shown the benefits of utilizing GP in SLAM, especially when the range data is sparse. However, they produce point-cloud maps without exporting the surfaces model. In addition, they cannot achieve real-time performance when using sensors with denser point clouds, like 64-beam LiDAR. This work exploits GP for mesh map building and is an upgraded version upon GP-SLAM+ \cite{ruan2020gp}. The main contributions of this work are summarized as follows:
\begin{itemize}
	\item We propose a novel meshing strategy based on GP reconstruction and vertices connection, which allows fast building, query, and updating of mesh maps.
	\item We design a point-to-mesh registration method. Together with constraints combination and multi-thread implementation, we ensure the efficiency and accuracy of localization and mapping with the mesh map.
	\item We develop a dense, real-time, and scalable open-source LiDAR SLAM system upon mesh map and demonstrate the advantages through experiments.
\end{itemize}

\section{Related Work}

Indoor dense mapping is more investigated than the outdoor case. A widely-adopted idea to build mesh is fusing depth data into a TSDF map and then extracting the implicit mesh on demand through the Marching cubes algorithm \cite{lorensen1987marching} like Voxblox \cite{oleynikova2017voxblox} and Kinectfusion \cite{newcombe2011kinectfusion}. On the contrary, some methods produce mesh maps explicitly. Ryde \textit{et al.} \cite{ryde2013voxel} voxelized point clouds and regressed a plane inside each voxel to form polygons. Piazza \textit{et al.} \cite{piazza2018real} used the Delaunay triangulation \cite{lee1980two} on the sparse feature points to build and update the mesh. Schreiberhuber \textit{et al.} \cite{schreiberhuber2019scalablefusion} connected the depth point in the camera image and then stitched the local mesh patches together. Our work explicitly produces mesh maps based on GP. Different from \cite{schreiberhuber2019scalablefusion}, our SLAMesh tackles unordered point clouds and builds mesh maps in the world frame rather than merging mesh patches in different image frames.

Recently, reliable 3-D LiDARs have enabled outdoor mapping. The outdoor environments are more challenging than indoor ones because the scenarios are several magnitudes larger, vehicle movements are fast, and the point cloud is sparse in remote areas.
Suma \cite{behley2018efficient} and Suma++ \cite{chen2019suma++} provide accurate odometry, but the surfel map is relatively cluttered. 
LiTAMIN2 \cite{yokozuka2021litamin2} shows the efficiency advantage of NDT registration with a map consisting of ellipsoids. 
Based on the TSDF map, K{\"u}hner \textit{et al.} \cite{kuhner2020large} modified the ray-tracking model for LiDAR and included an offline refinement process. 
Rold{\~a}o \textit{et al.} \cite{roldao20193d} combined explicit and implicit meshing approaches for better accuracy and mesh density. Those methods \cite{kuhner2020large}\cite{roldao20193d}  mainly focus on mapping function and can not provide real-time odometry. 
Puma \cite{vizzo2021poisson} and our SLAMesh both uses mesh maps. However, Puma utilizes ray-casting for registration and needs to rebuild the mesh map repetitively via the Poisson reconstruction \cite{kazhdan2006poisson} rather than updating it. Lin \textit{et al.} \cite{lin2023immesh} also investigated online meshing in LiDAR SLAM. They use a separate SLAM system and a direct connecting meshing strategy upon point-cloud maps. Our method reconstructs unordered points to integrate the mesh map into the whole system.

The ray-tracking model in the above TSDF-based methods neglects the fact that local surfaces are continuous. Conversely, GP builds a continuous model accompanied by uncertainty. This property attracts people to use GP for recovering occupancy field \cite{Kim2015} and TSDF field \cite{lee2019online}, which are continuous naturally. Different from \cite{Kim2015}\cite{lee2019online}, we use GP to predict the Euclidean coordinate of surfaces. Moreover, our system can also apply to large-scale environments.

\section{The Proposed Approach}

\begin{figure*}[t]
\setlength{\abovecaptionskip}{0pt} 
\setlength{\belowcaptionskip}{0pt} 
	\centering
	\includegraphics[scale = 0.45]{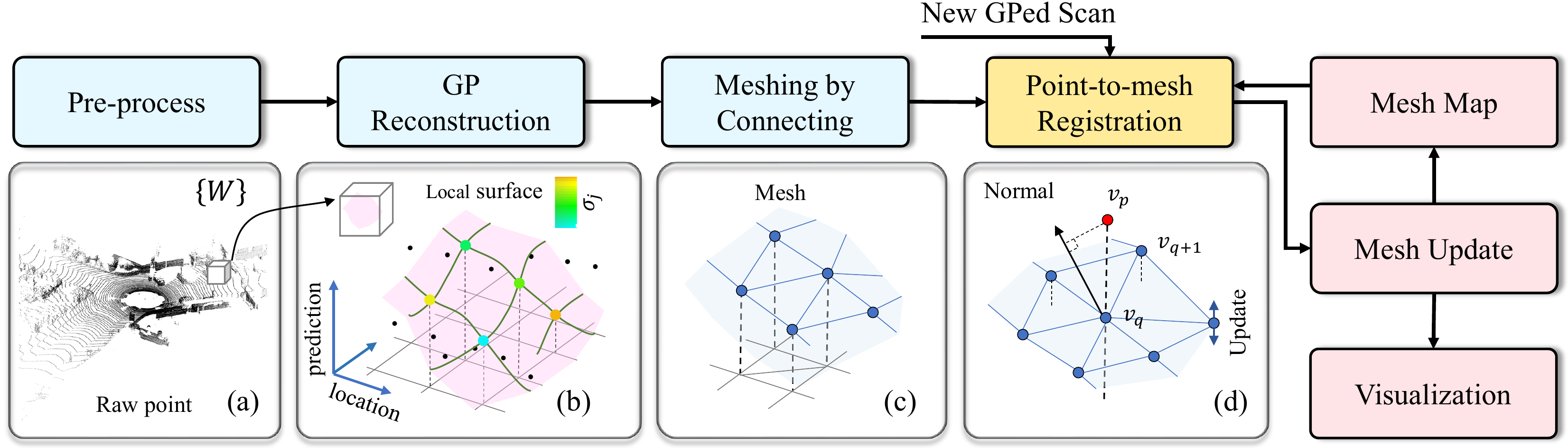}
	\caption{The general diagram of our SLAMesh. The system has three parts: meshing, registration, and mesh management. (a) The raw point cloud is transformed into the world frame $\{W\}$, down-sampled, and assigned into voxel cells; (b) GP models the local surface (light purple) of each cell to get uniformly distributed vertices. The color of vertices indicates their uncertainty (warmer means higher), which increases with its distance to the raw points; (c) We build mesh (light blue) by connecting adjacent vertices directly; (d) Point-to-mesh registration aligns the reconstructed current scan (red points) to the mesh map. Fast matching is feasible based on the \textit{locations} of vertices. The normal is smoothed from the surrounding mesh. GPed scan means scan after the GP reconstruction. Finally, we only need to adjust the 1-D \textit{predictions} of vertices and their uncertainty to maintain the mesh map. }
	\label{figworkflow}
	\vspace{-0.4cm}
\end{figure*} 

\subsection{Approach Overview}

The motivation of this work is to build, utilize, and maintain a mesh map in a LiDAR SLAM system. Fig. \ref{figworkflow} illustrates a general overview. The system mainly consists of three components: meshing, registration, and mesh management. Firstly, each new LiDAR scan is transformed into the world frame $\{W\}$ as $\mathcal{S}^{raw}$ using the initial guess from a constant-velocity model. The following operations are also performed in $\{W\}$. Then, points are assigned into voxel cells. GP reconstructs the local surface inside each cell and obtains vertices $\boldsymbol{v}_i$, which are connected to form a mesh. In the registration component, a point-to-mesh registration is designed to align the reconstructed current scan with the built mesh map $\mathcal{M}$. Finally, the mesh map is updated iteratively.
 
\subsection{Meshing Strategy}

As aforementioned, building and updating mesh is time-consuming. To tackle this problem, we adopt a reconstructing and connection strategy to facilitate the following pipeline so that the whole system can run in real-time. As shown in Fig. \ref{figworkflow},  GP recovers the local surface from noisy and sparse point clouds inside voxels. Then, the vertices are interpolation results of the surface. Two coordinates of the 3-D vertices are evenly located (named as \textit{locations}), and the other one (named as \textit{predictions}) has a continuous value domain. The \textit{locations} serve as the indices to enable fast query in constant time. The value domain of \textit{predictions} is continuous to avoid discretization-induced accuracy loss.  

Here the GP process is described (a more detailed description of GP can be found in \cite{li2020gp}\cite{ruan2020gp}). The bold lowercase letters represent vectors, and the uppercase letters represent matrices. The input to GP is a subset of the raw point cloud $\mathcal{S}_{k}^{raw} = \left\{\boldsymbol{p}_i = (x_i, y_i, z_i, \sigma_{in}^{2}), i=1,....,n_i \right\}$ containing $n_i$ points, where $k$ represents the $k$-th cell $\mathcal{C}_k$ in the current scan, $\sigma_{in}^{2}$ is the isotropic variance of input noise. The output is a \textit{layer} containing $n_j$ vertices $\mathcal{L}_k^{z} = \left\{\boldsymbol{v}_j=(x_j, y_j, z_j, \sigma_j^{2}), j=1,....,n_j\right\}$ with uncertainty $\sigma_j^{2}$. The superscript $z$ means that GP predicts the coordinate $z$ as $\boldsymbol{ \widetilde{f}} = \left\{z_j, j=1,....,n_j\right\}$, and it is omitted when the coordinate can be any one in $x$, $y$, or $z$. We denote the \textit{locations} of input and output point set as $\boldsymbol{i}$ and $\boldsymbol{j}$. Given the input observation $\boldsymbol{f}$, the \textit{predictions} $\boldsymbol{ \widetilde{f}}$ also follow the Gaussian distribution. The expectation of $\boldsymbol{ \widetilde{f}}$ is:
\begin{equation}
\boldsymbol{ \widetilde{f}}=\boldsymbol{k}_{\boldsymbol{ij}}^{T}\left(\sigma_{in}^{2} \boldsymbol{I}+\boldsymbol{K}_{\boldsymbol{ii}}\right)^{-1} \boldsymbol{f}.
\end{equation}
The uncertainty of \textit{predictions} is its variance:
\begin{equation}
\boldsymbol{\sigma_j}^{2} = \boldsymbol{k}_{\boldsymbol{jj}}-\boldsymbol{k}_{\boldsymbol{ij}}^{\boldsymbol{T}}\left(\sigma_{in}^{2} \boldsymbol{I}+\boldsymbol{K}_{\boldsymbol{ii}}\right)^{-1} \boldsymbol{k}_{\boldsymbol{ij}},
\end{equation}
where $\boldsymbol{k}_{\boldsymbol{jj}}$, $\boldsymbol{k}_{\boldsymbol{ij}}$, and $\boldsymbol{K}_{\boldsymbol{i} \boldsymbol{i}}$ represent different combinations of the kernel function of \textit{locations}. For instance, 
\begin{equation}
k\left(i,j\right)=\exp \left(-\kappa\left|i-j\right|\right),
\end{equation}
where $k\left(i,j\right)$ is a scale value, and $\kappa$ is a constant ($\kappa=1$ in our algorithm). This exponential kernel function can represent a local smooth surface in a 2-D manifold. Thus, one cell can  contains more mesh \textit{layers} with different GP functions (up to $3$ if all coordinates are interpolated) when modeling a complicated structure \cite{ruan2020gp}. 

A vertex with uncertainty $\sigma_j^2$ under a certain threshold $\sigma_{match}^2$ is accurate enough or valid. Triangle mesh faces are built by connecting adjacent or diagonal vertices in the 2-D space of \textit{locations}. A mesh face is valid if all vertices are valid. Similar to the Delaunay triangulation, which establishes edges in a 2-D space, our method can prevent sliver mesh faces. 
 
\subsection{Point-to-Mesh Registration}

As aforementioned, we perform localization and meshing simultaneously so that meshing can benefit localization. To this end, one intuitive idea is to treat vertices as points or extract points from faces and then employ traditional point-cloud registration for pose estimation. However, this idea neglects the normal information of the mesh faces. Puma \cite{vizzo2021poisson} indicates that a point-to-mesh error could improve accuracy. Unlike the ray-casting data association in Puma, our SLAMesh establishes correspondences based on \textit{locations}.
 
\begin{figure}[t]
\setlength{\abovecaptionskip}{0pt} 
\setlength{\belowcaptionskip}{0pt} 
	\centering
	\includegraphics[scale = 0.38]{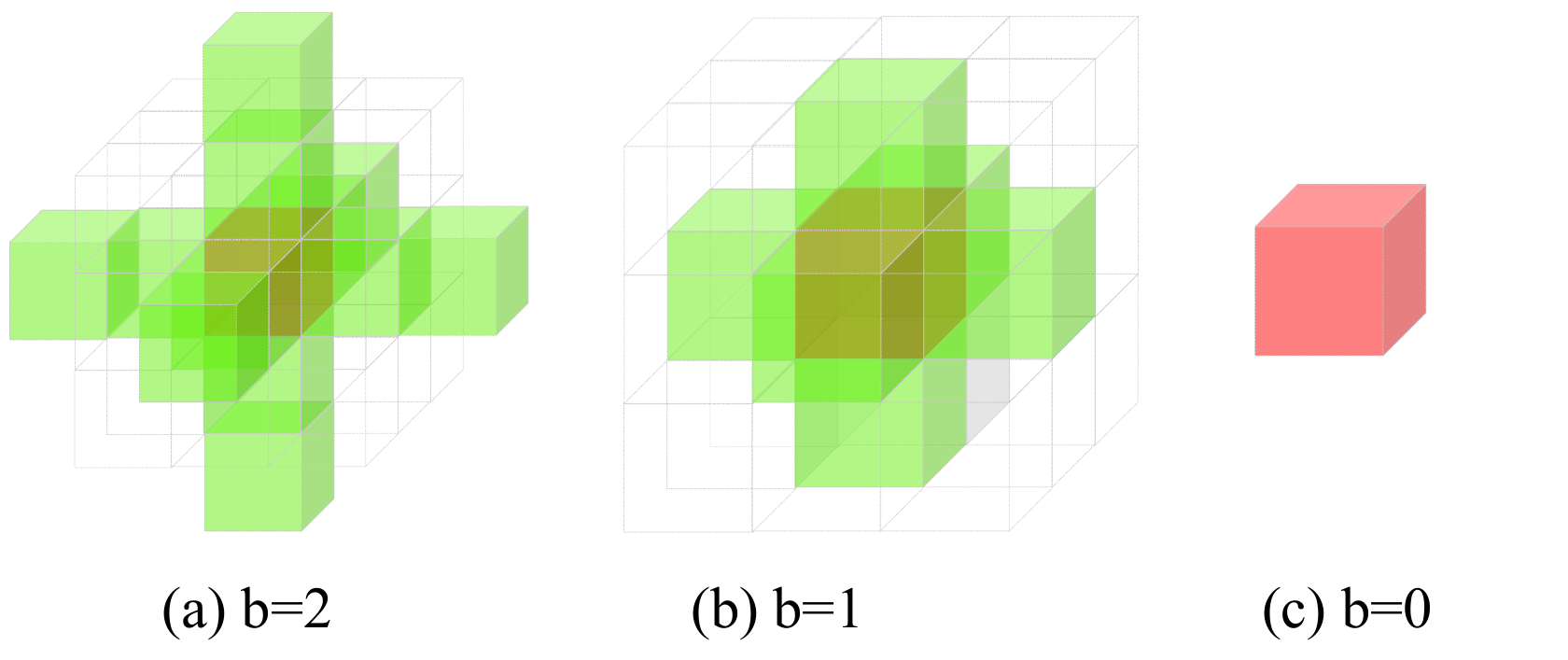}
	\caption{The data association can be established across cells alongside the predicted coordinates (green cells). The query length $b$ can be decreased as registration tends to converge. From (a) to (c), $b$ is decreased from $2$ to $0$.}
	\label{figoverlap}
    \vspace{-0.6cm}
\end{figure}

For a vertex $\boldsymbol{v}_p$ in the reconstructed current scan $\mathcal{S}^{gp}$, we first query the sub-mesh \textit{layer} located in the same or adjacent cells (see Fig. \ref{figoverlap}), and then find the vertex $\boldsymbol{v}_q$ sharing the same \textit{locations} (see Fig. \ref{figworkflow}(d)). Data association is established between $\boldsymbol{v}_p$ and the valid faces that contain $\boldsymbol{v}_q$. Finding the $n_q$ neighboring vertices along the edges in a mesh is quick. The normals of those faces that contain $\boldsymbol{v}_q$ are averaged to get a smoothed normal $\overline{\boldsymbol{n}}$ in case the surface is rugged:
\begin{equation}
\overline{\boldsymbol{n}_q}=  \frac{ \sum_{q=1}^{n_q-2} (\boldsymbol{v}_{q}- \boldsymbol{v}_{q+1})\times(\boldsymbol{v}_{q} - \boldsymbol{v}_{q+2}) }
{ \sum_{q=1}^{n_q-2} \left\|(\boldsymbol{v}_{q}- \boldsymbol{v}_{q+1})\times(\boldsymbol{v}_{q} - \boldsymbol{v}_{q+2}) \right\|},
\end{equation}
where $\left\|\cdot\right\|$ is the 2-norm. We denote the number of correspondences in one cell as $n_q$ and the number of overlapped cells as $K$. The point $\boldsymbol{v}_p^{\prime}$ and the point-to-mesh residual after applying the transformation $\boldsymbol{T}\in SE3$ including rotation $\boldsymbol{R}$ and translation $\boldsymbol{t}$ on $\boldsymbol{v}_p$ are:
\begin{equation}
\left[                
  \begin{array}{c}   
    \boldsymbol{v}_p^{\prime}\\ 
     1\\ 
  \end{array}
\right]
 = \boldsymbol{T} \left[                
  \begin{array}{c}   
    \boldsymbol{v}_p\\ 
     1\\ 
  \end{array}
\right] = 
\left[                
  \begin{array}{cc}   
    \boldsymbol{R} & \boldsymbol{t}\\  
    \boldsymbol{0} & 1\\  
  \end{array}
\right] 
\left[                
  \begin{array}{c}   
    \boldsymbol{v}_p\\ 
     1\\ 
  \end{array}
\right],
\end{equation}
\begin{equation}
e_p=\overline{\boldsymbol{n}_q^{T}} (\boldsymbol{v}_p^{\prime} - \boldsymbol{v}_q ), 
\end{equation}
where the superscript $T$ means matrix transpose. The optimal relative transformation $\boldsymbol{T}$ is computed in the optimization problem where point-to-mesh residuals in all $K$ overlapped \textit{layers} $\mathcal{L}_k$ are joined:
\begin{equation}
{\boldsymbol{T}}=\underset{\boldsymbol{T}}{{argmin}} \sum_{k=1}^{K} \sum_{\boldsymbol{v}_p \in \mathcal{L}_k,p=1}^{n_p} e_p.
\end{equation}
We use the Levenberg–Marquardt algorithm in Ceres solver$\footnote{\href{http://ceres-solver.org/}{http://ceres-solver.org/}}$ to solve this non-linear least square problem. The analytical Jacobin matrix is derived to accelerate the solving process: 
\begin{equation}
\boldsymbol{J}=\frac{\partial e_p}{\partial \boldsymbol{T}}=\frac{\partial e_p}{\partial \boldsymbol{v}_p^{\prime}} \frac{\partial \boldsymbol{v}_p^{\prime}}{\partial \boldsymbol{T}},
\end{equation}
\begin{equation}
\frac{\partial e_p}{\partial \boldsymbol{v}_p^{\prime}}=\overline{\boldsymbol{n}_p^{T}},
\hspace{1.5em}
\frac{\partial \boldsymbol{v}_p^{\prime}}{\partial \boldsymbol{t}}=\boldsymbol{I},
\hspace{1.5em}
  \frac{\partial \boldsymbol{v}_p^{\prime}}{\partial \boldsymbol{R}}=-\left(\boldsymbol{R} \boldsymbol{v}_p+\boldsymbol{t}\right)_{\times},
\end{equation}
where the symbol $\left(\cdot\right)_{\times}$ represents the corresponding skew-symmetric matrix of a vector. The mesh may contain many faces so the optimization problem can be huge. We combine the residuals inside each \textit{layer} to be one by averaging before optimization to speed up the optimization process. 
  
Storing data in voxel cells induces discontinuities of data association on the border. If we only perform registration on overlapped cells, when the vehicle moves fast, the convergence basins of registration would be too narrow. Thus, we allow cross-cell overlapping by querying cells alongside the axis of \textit{predictions} for each \textit{layer} as shown in Fig. \ref{figoverlap}. The length of the query is decreased when the registration tends to converge.

\subsection{Mesh Management and Multithreading}

Mesh maps are more challenging to maintain than point-cloud maps and grid-based maps. Points and grids can be independent of their neighborhoods. On the contrary, mesh retains the topological structure with vertices connected to each other. The map update process should maintain existing connections and avoid sliver triangles. Typical solutions include re-meshing with Delaunay triangulation \cite{piazza2018real}, maintaining an implicit filed similar to Voxblox \cite{oleynikova2017voxblox}, or repeating the meshing process like Puma \cite{vizzo2021poisson}. SLAM requires a fast update of mesh. In our SLAMesh, the \textit{locations} of vertices are fixed, so only the 1-D \textit{predictions} need to be updated. Given the previous $t$ data below the threshold $\sigma_{update}^2$, the new \textit{prediction} can be solved by the iterative least square:
\begin{equation}
{ \widetilde{f}_{t_k}}={ \sum_{t=1}^{t_k} ({ \widetilde{f}_t} \sigma_{t}^{2}) }/{ \sum_{t=1}^{t_k} (\sigma_{t}^{2}), \text{if $\sigma_{t}^{2} < \sigma_{update}^2$}}.
\end{equation}

The cells are stored in a hash map, which ensures a linear complexity for inserting, deleting, and querying. Also, this structure is flexible because it can grow incrementally. An advantage of our cell-based map shared with VGICP \cite{koide2021voxelized} is the compatibility for parallel processing. Since each cell is independent in these steps, we can accelerate the GP reconstruction and mesh publishing processes with multithreading. 

\section{Experimental Results and Discussions}

\subsection{Experimental Setup}
We compare SLAMesh with several state-of-the-art SLAM systems or meshing tools. Algorithms are run on a PC with a 3.6 GHz 8-core Intel i7-11700KF CPU. In SLAMesh, the number of \textit{predictions} in each cell $n_j =6^{2}$. The cell size is $1.5$m so that the resolution of \textit{locations} is identical to the voxel size in Voxblox when evaluating mesh. In other tests, the cell size is $1.6$m. The variance of the input noise is $\sigma_{in}^2 = 0.02$. The thresholds $\sigma_{match}^2 = 0.5$ and $\sigma_{update}^2 = 1.0 $. These parameters are tuned manually. $8$ threads are allocated in multithreading. We use the Mesh Tools \cite{putz2021mesh} for visualization. Other methods are run in their default configurations. 

\begin{figure}[tbp]
\setlength{\abovecaptionskip}{0pt} 
\setlength{\belowcaptionskip}{0pt} 
	\centering
	\includegraphics[scale = 0.35]{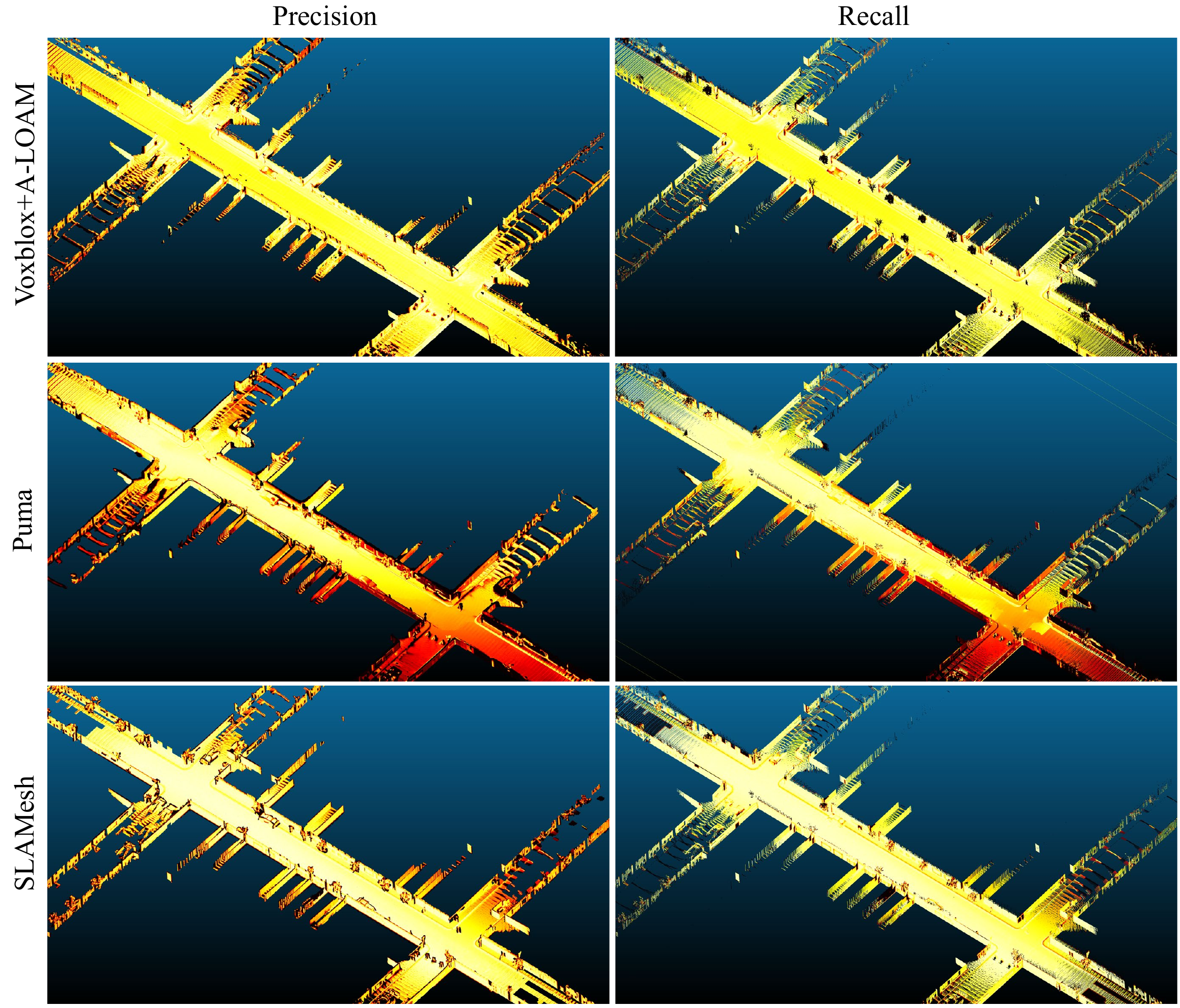}
	\caption{Mesh maps compared with the ground-truth model on the Mai City dataset \cite{vizzo2021poisson}. The left and right columns illustrate Precision and Recall values, respectively. Darker colors indicate worse performance and vice versa. Voxblox recovers good results but fails to build thin structures like trees. Puma presents less accuracy on the edges of structures or areas where the pose drifts. Our SLAMesh presents both good Recall and Precision.} 
	\label{figmeshkitti}
 	\vspace{-0.2cm}
\end{figure}

\subsection{Meshing Evaluation}

The first experiment evaluates the meshing quality in terms of accuracy and completeness. We use the Mai City dataset \cite{vizzo2021poisson} built on the CARLA simulation environment with ground truth of maps. In this dataset, the sensor is a simulated 64-beam Velodyne HDL-64E LiDAR. The vehicle drives for $99$m and produces $100$ frames of LiDAR scans. The ground truth of the environment is a dense point cloud scanned by a high-resolution sensor. We evaluate the combination of Voxblox with A-LOAM$\footnote{\href{https://github.com/HKUST-Aerial-Robotics/A-LOAM}{https://github.com/HKUST-Aerial-Robotics/A-LOAM}}$, Puma, and SLAMesh. Because this work aims to build a SLAM system rather than an offline mapping tool, and the ground truth pose is usually unavailable in real environments, we use the poses estimated in each pipeline or an extra odometry system (i.e., using A-LOAM for Voxblox).

Fig. \ref{figmeshsample} presents the mesh map produced by each method. All maps reconstruct the main structures well. In the zoomed views, we can observe the characteristic of each strategy. The vertices in SLAMesh and Voxblox are evenly organized due to the reconstruction or voxelization. The mesh of Puma is water-tight with Poisson reconstruction but is relatively complex with multi-layer phenomenons. We believe that this is because SLAMesh and Voxblox both iteratively fuse observation into surfaces, while Puma accumulates several scans and then conducts a meshing upon them. 
 
The quantitative evaluation uses the standard point-cloud-based metrics: Precision, Recall, and F1-score \cite{knapitsch2017tanks}. We densely and uniformly sample the mesh maps to form a point cloud, and then compare it with the ground-truth point cloud. The quantitative results are displayed in Table. \ref{tabmeshevaluation}, where the distance threshold $d$ is set to $0.3$m. The performance of our SLAMesh is best, followed by Voxblox+A-LOAM and Puma. Fig. \ref{figmeshkitti} visualizes the Precision and Recall results. The values decrease from lighter to darker. Voxblox erases tiny or thin objects like trees (the black areas in the Recall sub-figures) with its ray-tracing method. Puma accumulates errors along the trajectory. The water-tight assumption makes the top edge of walls slightly warped. Our SLAMesh can recover both large and small structures with high accuracy. 

\subsection{Odometry Evaluation}
 
\begin{table}[t]
\setlength{\abovecaptionskip}{0pt} 
\setlength{\belowcaptionskip}{0pt} 
\renewcommand\arraystretch{1.0} 
\renewcommand\tabcolsep{10pt}	
   \renewcommand{\baselinestretch}{0.6}
   \caption{Quantitative comparative results ($\%$) of meshing. The bold font indicates best results.}
 \centering
	\label{tabmeshevaluation}
	\begin{tabular}{@{}c c c c@{}}
		\toprule[1.3pt] 
		Metric 		& Voxblox\cite{oleynikova2017voxblox}+A-LOAM   & Puma \cite{vizzo2021poisson} & SLAMesh (Ours)\\
		\toprule[0.5pt] 
		Precision & 69.50  & 43.00  & \textbf{74.96}  \\
		Recall    & 85.91  & 62.90  & \textbf{86.09}  \\
		F1-score   & 76.83  & 51.08  & \textbf{80.14}  \\
		\toprule[1.3pt] 
	\end{tabular}
\end{table}

\renewcommand\arraystretch{0.7}
\newcommand{\tabincell}[2]{\begin{tabular}{@{}#1@{}}#2\end{tabular}}  

\begin{table*}[t]
\setlength{\abovecaptionskip}{0pt} 
\setlength{\belowcaptionskip}{0pt} 
\renewcommand\arraystretch{0.5} 
\renewcommand\tabcolsep{6.2pt}
  \centering
  \renewcommand{\baselinestretch}{0.8}
  \caption{Quantitative results of odometry accuracy on the KITTI dataset. The first and the second row are the Relative Translation Error (\%) and Rotation Error (deg/100m), respectively. The red bold and the blue fonts indicate the best and the second best results, respectively. Comb. means constraints combination, and P2Mesh means point-to-mesh registration.}
  \centering
 	\label{tab2kittiresult}
    \begin{tabular}{ccccccccccccccc}
    \toprule
    \multicolumn{1}{l}{Map type} & \multicolumn{2}{c}{Method} & 00    & 01    & 02    & 03    & 04    & 05    & 06    & 07    & 08    & 09    & 10    & Mean \\
    \midrule
    \multirow{4}[4]{*}{Point Cloud} & \multicolumn{2}{c}{\multirow{2}[2]{*}{LOAM \cite{zhang2017low}}} & \multicolumn{1}{r}{\cellcolor[rgb]{ .949,  .949,  .949} 0.78} & \multicolumn{1}{r}{\cellcolor[rgb]{ .949,  .949,  .949} 1.43} & \multicolumn{1}{r}{\cellcolor[rgb]{ .949,  .949,  .949} 0.92} & \multicolumn{1}{r}{\cellcolor[rgb]{ .949,  .949,  .949} 0.86} & \multicolumn{1}{r}{\cellcolor[rgb]{ .949,  .949,  .949} 0.71} & \multicolumn{1}{r}{\cellcolor[rgb]{ .949,  .949,  .949} 0.57} & \multicolumn{1}{r}{\cellcolor[rgb]{ .949,  .949,  .949} 0.65} & \multicolumn{1}{r}{\cellcolor[rgb]{ .949,  .949,  .949} 0.63} & \multicolumn{1}{r}{\cellcolor[rgb]{ .949,  .949,  .949} 1.12} & \multicolumn{1}{r}{\cellcolor[rgb]{ .949,  .949,  .949} 0.77} & \multicolumn{1}{r}{\cellcolor[rgb]{ .949,  .949,  .949} 0.79} & \multicolumn{1}{r}{\cellcolor[rgb]{ .949,  .949,  .949} 0.839} \\
          & \multicolumn{2}{c}{} & \multicolumn{1}{c}{-} & \multicolumn{1}{c}{-} & \multicolumn{1}{c}{-} & \multicolumn{1}{c}{-} & \multicolumn{1}{c}{-} & \multicolumn{1}{c}{-} & \multicolumn{1}{c}{-} & \multicolumn{1}{c}{-} & \multicolumn{1}{c}{-} & \multicolumn{1}{c}{-} & \multicolumn{1}{c}{-} & \multicolumn{1}{c}{-} \\
\cmidrule{2-15}          & \multicolumn{2}{c}{\multirow{2}[2]{*}{A-LOAM}} & \multicolumn{1}{r}{\cellcolor[rgb]{ .949,  .949,  .949} 0.97} & \multicolumn{1}{r}{\cellcolor[rgb]{ .949,  .949,  .949} 2.75} & \multicolumn{1}{r}{\cellcolor[rgb]{ .949,  .949,  .949} 4.91} & \multicolumn{1}{r}{\cellcolor[rgb]{ .949,  .949,  .949} 1.22} & \multicolumn{1}{r}{\cellcolor[rgb]{ .949,  .949,  .949} 1.35} & \multicolumn{1}{r}{\cellcolor[rgb]{ .949,  .949,  .949} 0.63} & \multicolumn{1}{r}{\cellcolor[rgb]{ .949,  .949,  .949} 0.61} & \multicolumn{1}{r}{\cellcolor[rgb]{ .949,  .949,  .949} 0.48} & \multicolumn{1}{r}{\cellcolor[rgb]{ .949,  .949,  .949} 1.17} & \multicolumn{1}{r}{\cellcolor[rgb]{ .949,  .949,  .949} 1.11} & \multicolumn{1}{r}{\cellcolor[rgb]{ .949,  .949,  .949} 1.58} & \multicolumn{1}{r}{\cellcolor[rgb]{ .949,  .949,  .949} 1.525} \\
          & \multicolumn{2}{c}{} & \multicolumn{1}{r}{0.74} & \multicolumn{1}{r}{0.96} & \multicolumn{1}{r}{2.70} & \multicolumn{1}{r}{1.06} & \multicolumn{1}{r}{0.67} & \multicolumn{1}{r}{0.51} & \multicolumn{1}{r}{0.59} & \multicolumn{1}{r}{0.66} & \multicolumn{1}{r}{0.76} & \multicolumn{1}{r}{0.75} & \multicolumn{1}{r}{0.82} & \multicolumn{1}{r}{0.928} \\
    \midrule
    \midrule
    \multirow{4}[4]{*}{Surfel} & \multicolumn{2}{c}{\multirow{2}[2]{*}{Suma \cite{behley2018efficient}}} & \multicolumn{1}{r}{\cellcolor[rgb]{ .949,  .949,  .949} \textcolor[rgb]{ 0,  .439,  .753}{0.70}} & \multicolumn{1}{r}{\cellcolor[rgb]{ .949,  .949,  .949} 1.70} & \multicolumn{1}{r}{\cellcolor[rgb]{ .949,  .949,  .949} 1.10} & \multicolumn{1}{r}{\cellcolor[rgb]{ .949,  .949,  .949} 0.70} & \multicolumn{1}{r}{\cellcolor[rgb]{ .949,  .949,  .949} \textcolor[rgb]{ 0,  .439,  .753}{0.40}} & \multicolumn{1}{r}{\cellcolor[rgb]{ .949,  .949,  .949} 0.50} & \multicolumn{1}{r}{\cellcolor[rgb]{ .949,  .949,  .949} \textcolor[rgb]{ 0,  .439,  .753}{0.40}} & \multicolumn{1}{r}{\cellcolor[rgb]{ .949,  .949,  .949} 0.40} & \multicolumn{1}{r}{\cellcolor[rgb]{ .949,  .949,  .949} 1.00} & \multicolumn{1}{r}{\cellcolor[rgb]{ .949,  .949,  .949} \textcolor[rgb]{ 0,  .439,  .753}{0.50}} & \multicolumn{1}{r}{\cellcolor[rgb]{ .949,  .949,  .949} 0.70} & \multicolumn{1}{r}{\cellcolor[rgb]{ .949,  .949,  .949} 0.736} \\
          & \multicolumn{2}{c}{} & \multicolumn{1}{r}{0.30} & \multicolumn{1}{r}{0.30} & \multicolumn{1}{r}{0.40} & \multicolumn{1}{r}{0.50} & \multicolumn{1}{r}{0.30} & \multicolumn{1}{r}{\textcolor[rgb]{ 1,  0,  0}{\textbf{0.20}}} & \multicolumn{1}{r}{\textcolor[rgb]{ 0,  .439,  .753}{0.20}} & \multicolumn{1}{r}{\textcolor[rgb]{ 1,  0,  0}{\textbf{0.30}}} & \multicolumn{1}{r}{0.40} & \multicolumn{1}{r}{0.30} & \multicolumn{1}{r}{\textcolor[rgb]{ 0,  .439,  .753}{0.30}} & \multicolumn{1}{r}{0.318} \\
\cmidrule{2-15}          & \multicolumn{2}{c}{\multirow{2}[2]{*}{Suma++ \cite{chen2019suma++}}} & \multicolumn{1}{r}{\cellcolor[rgb]{ .949,  .949,  .949} \textcolor[rgb]{ 1,  0,  0}{\textbf{0.64}}} & \multicolumn{1}{r}{\cellcolor[rgb]{ .949,  .949,  .949} 1.60} & \multicolumn{1}{r}{\cellcolor[rgb]{ .949,  .949,  .949} 1.00} & \multicolumn{1}{r}{\cellcolor[rgb]{ .949,  .949,  .949} 0.67} & \multicolumn{1}{r}{\cellcolor[rgb]{ .949,  .949,  .949} \textcolor[rgb]{ 1,  0,  0}{\textbf{0.37}}} & \multicolumn{1}{r}{\cellcolor[rgb]{ .949,  .949,  .949} \textcolor[rgb]{ 1,  0,  0}{\textbf{0.40}}} & \multicolumn{1}{r}{\cellcolor[rgb]{ .949,  .949,  .949} 0.46} & \multicolumn{1}{r}{\cellcolor[rgb]{ .949,  .949,  .949} \textcolor[rgb]{ 1,  0,  0}{\textbf{0.34}}} & \multicolumn{1}{r}{\cellcolor[rgb]{ .949,  .949,  .949} 1.10} & \multicolumn{1}{r}{\cellcolor[rgb]{ .949,  .949,  .949} \textcolor[rgb]{ 1,  0,  0}{\textbf{0.47}}} & \multicolumn{1}{r}{\cellcolor[rgb]{ .949,  .949,  .949} \textcolor[rgb]{ 0,  .439,  .753}{0.66}} & \multicolumn{1}{r}{\cellcolor[rgb]{ .949,  .949,  .949} \textcolor[rgb]{ 0,  .439,  .753}{0.701}} \\
          & \multicolumn{2}{c}{} & \multicolumn{1}{r}{\textcolor[rgb]{ 1,  0,  0}{\textbf{0.22}}} & \multicolumn{1}{r}{0.46} & \multicolumn{1}{r}{0.37} & \multicolumn{1}{r}{0.46} & \multicolumn{1}{r}{\textcolor[rgb]{ 0,  .439,  .753}{0.26}} & \multicolumn{1}{r}{\textcolor[rgb]{ 1,  0,  0}{\textbf{0.20}}} & \multicolumn{1}{r}{0.21} & \multicolumn{1}{r}{0.19} & \multicolumn{1}{r}{0.35} & \multicolumn{1}{r}{\textcolor[rgb]{ 1,  0,  0}{\textbf{0.23}}} & \multicolumn{1}{r}{\textcolor[rgb]{ 1,  0,  0}{\textbf{0.28}}} & \multicolumn{1}{r}{\textcolor[rgb]{ 0,  .439,  .753}{0.294}} \\
    \midrule
    \midrule
    \multirow{2}[2]{*}{NDT} & \multicolumn{2}{c}{\multirow{2}[2]{*}{Litamin2 \cite{yokozuka2021litamin2}}} & \multicolumn{1}{r}{\cellcolor[rgb]{ .949,  .949,  .949} 0.70} & \multicolumn{1}{r}{\cellcolor[rgb]{ .949,  .949,  .949} 2.10} & \multicolumn{1}{r}{\cellcolor[rgb]{ .949,  .949,  .949} 0.98} & \multicolumn{1}{r}{\cellcolor[rgb]{ .949,  .949,  .949} 0.96} & \multicolumn{1}{r}{\cellcolor[rgb]{ .949,  .949,  .949} 1.05} & \multicolumn{1}{r}{\cellcolor[rgb]{ .949,  .949,  .949} 0.45} & \multicolumn{1}{r}{\cellcolor[rgb]{ .949,  .949,  .949} 0.59} & \multicolumn{1}{r}{\cellcolor[rgb]{ .949,  .949,  .949} 0.44} & \multicolumn{1}{r}{\cellcolor[rgb]{ .949,  .949,  .949} 0.95} & \multicolumn{1}{r}{\cellcolor[rgb]{ .949,  .949,  .949} 0.69} & \multicolumn{1}{r}{\cellcolor[rgb]{ .949,  .949,  .949} 0.80} & \multicolumn{1}{r}{\cellcolor[rgb]{ .949,  .949,  .949} 0.883} \\
          & \multicolumn{2}{c}{} & \multicolumn{1}{r}{\textcolor[rgb]{ 0,  .439,  .753}{0.28}} & \multicolumn{1}{r}{0.46} & \multicolumn{1}{r}{\textcolor[rgb]{ 0,  .439,  .753}{0.32}} & \multicolumn{1}{r}{0.48} & \multicolumn{1}{r}{0.52} & \multicolumn{1}{r}{\textcolor[rgb]{ 0,  .439,  .753}{0.25}} & \multicolumn{1}{r}{0.34} & \multicolumn{1}{r}{0.32} & \multicolumn{1}{r}{0.29} & \multicolumn{1}{r}{0.40} & \multicolumn{1}{r}{0.47} & \multicolumn{1}{r}{0.375} \\
    \midrule
    \midrule
    \multirow{8}[8]{*}{Mesh} & \multicolumn{2}{c}{\multirow{2}[2]{*}{Puma \cite{vizzo2021poisson}}} & \multicolumn{1}{r}{\cellcolor[rgb]{ .949,  .949,  .949} 1.46} & \multicolumn{1}{r}{\cellcolor[rgb]{ .949,  .949,  .949} 3.38} & \multicolumn{1}{r}{\cellcolor[rgb]{ .949,  .949,  .949} 1.86} & \multicolumn{1}{r}{\cellcolor[rgb]{ .949,  .949,  .949} 1.60} & \multicolumn{1}{r}{\cellcolor[rgb]{ .949,  .949,  .949} 1.63} & \multicolumn{1}{r}{\cellcolor[rgb]{ .949,  .949,  .949} 1.20} & \multicolumn{1}{r}{\cellcolor[rgb]{ .949,  .949,  .949} 0.88} & \multicolumn{1}{r}{\cellcolor[rgb]{ .949,  .949,  .949} 0.72} & \multicolumn{1}{r}{\cellcolor[rgb]{ .949,  .949,  .949} 1.44} & \multicolumn{1}{r}{\cellcolor[rgb]{ .949,  .949,  .949} 1.51} & \multicolumn{1}{r}{\cellcolor[rgb]{ .949,  .949,  .949} 1.38} & \multicolumn{1}{r}{\cellcolor[rgb]{ .949,  .949,  .949} 1.551} \\
          & \multicolumn{2}{c}{} & \multicolumn{1}{r}{0.68} & \multicolumn{1}{r}{1.00} & \multicolumn{1}{r}{0.72} & \multicolumn{1}{r}{1.10} & \multicolumn{1}{r}{0.92} & \multicolumn{1}{r}{0.61} & \multicolumn{1}{r}{0.42} & \multicolumn{1}{r}{0.55} & \multicolumn{1}{r}{0.61} & \multicolumn{1}{r}{0.66} & \multicolumn{1}{r}{0.84} & \multicolumn{1}{r}{0.737} \\
\cmidrule{2-15}          & \multirow{6}[6]{*}{\tabincell{c}{SLAMesh \\(Ours)}} & \multirow{2}[2]{*}{Full} & \multicolumn{1}{r}{\cellcolor[rgb]{ .949,  .949,  .949} 0.77} & \multicolumn{1}{r}{\cellcolor[rgb]{ .949,  .949,  .949} \textcolor[rgb]{ 1,  0,  0}{\textbf{1.25}}} & \multicolumn{1}{r}{\cellcolor[rgb]{ .949,  .949,  .949} \textcolor[rgb]{ 1,  0,  0}{\textbf{0.77}}} & \multicolumn{1}{r}{\cellcolor[rgb]{ .949,  .949,  .949} \textcolor[rgb]{ 0,  .439,  .753}{0.64}} & \multicolumn{1}{r}{\cellcolor[rgb]{ .949,  .949,  .949} 0.50} & \multicolumn{1}{r}{\cellcolor[rgb]{ .949,  .949,  .949} 0.52} & \multicolumn{1}{r}{\cellcolor[rgb]{ .949,  .949,  .949} 0.53} & \multicolumn{1}{r}{\cellcolor[rgb]{ .949,  .949,  .949} \textcolor[rgb]{ 0,  .439,  .753}{0.36}} & \multicolumn{1}{r}{\cellcolor[rgb]{ .949,  .949,  .949} \textcolor[rgb]{ 0,  .439,  .753}{0.87}} & \multicolumn{1}{r}{\cellcolor[rgb]{ .949,  .949,  .949} 0.57} & \multicolumn{1}{r}{\cellcolor[rgb]{ .949,  .949,  .949} \textcolor[rgb]{ 1,  0,  0}{\textbf{0.65}}} & \multicolumn{1}{r}{\cellcolor[rgb]{ .949,  .949,  .949} \textcolor[rgb]{ 1,  0,  0}{\textbf{0.676}}} \\
          &       &       & \multicolumn{1}{r}{0.35} & \multicolumn{1}{r}{\textcolor[rgb]{ 0,  .439,  .753}{0.30}} & \multicolumn{1}{r}{\textcolor[rgb]{ 1,  0,  0}{\textbf{0.30}}} & \multicolumn{1}{r}{\textcolor[rgb]{ 0,  .439,  .753}{0.43}} & \multicolumn{1}{r}{\textcolor[rgb]{ 1,  0,  0}{\textbf{0.13}}} & \multicolumn{1}{r}{0.30} & \multicolumn{1}{r}{0.22} & \multicolumn{1}{r}{\textcolor[rgb]{ 0,  .439,  .753}{0.23}} & \multicolumn{1}{r}{\textcolor[rgb]{ 1,  0,  0}{\textbf{0.27}}} & \multicolumn{1}{r}{\textcolor[rgb]{ 0,  .439,  .753}{0.25}} & \multicolumn{1}{r}{0.42} & \multicolumn{1}{r}{\textcolor[rgb]{ 1,  0,  0}{\textbf{0.291}}} \\
\cmidrule{3-15}          &       & \multirow{2}[2]{*}{\tabincell{c}{w/o Comb.}} & \multicolumn{1}{r}{\cellcolor[rgb]{ .949,  .949,  .949} 0.96} & \multicolumn{1}{r}{\cellcolor[rgb]{ .949,  .949,  .949} 1.70} & \multicolumn{1}{r}{\cellcolor[rgb]{ .949,  .949,  .949} 1.04} & \multicolumn{1}{r}{\cellcolor[rgb]{ .949,  .949,  .949} 0.81} & \multicolumn{1}{r}{\cellcolor[rgb]{ .949,  .949,  .949} 0.67} & \multicolumn{1}{r}{\cellcolor[rgb]{ .949,  .949,  .949} 0.51} & \multicolumn{1}{r}{\cellcolor[rgb]{ .949,  .949,  .949} 0.64} & \multicolumn{1}{r}{\cellcolor[rgb]{ .949,  .949,  .949} 0.46} & \multicolumn{1}{r}{\cellcolor[rgb]{ .949,  .949,  .949} \textcolor[rgb]{ 1,  0,  0}{\textbf{0.86}}} & \multicolumn{1}{r}{\cellcolor[rgb]{ .949,  .949,  .949} 0.74} & \multicolumn{1}{r}{\cellcolor[rgb]{ .949,  .949,  .949} 1.02} & \multicolumn{1}{r}{\cellcolor[rgb]{ .949,  .949,  .949} 0.857} \\
          &       &       & \multicolumn{1}{r}{0.41} & \multicolumn{1}{r}{0.41} & \multicolumn{1}{r}{0.39} & \multicolumn{1}{r}{0.55} & \multicolumn{1}{r}{0.30} & \multicolumn{1}{r}{0.29} & \multicolumn{1}{r}{0.30} & \multicolumn{1}{r}{0.28} & \multicolumn{1}{r}{\textcolor[rgb]{ 0,  .439,  .753}{0.30}} & \multicolumn{1}{r}{0.32} & \multicolumn{1}{r}{0.44} & \multicolumn{1}{r}{0.363} \\
\cmidrule{3-15}          &       & \multirow{2}[2]{*}{\tabincell{c}{w/o P2Mesh}} & \multicolumn{1}{r}{\cellcolor[rgb]{ .949,  .949,  .949} 0.70} & \multicolumn{1}{r}{\cellcolor[rgb]{ .949,  .949,  .949} \textcolor[rgb]{ 0,  .439,  .753}{1.33}} & \multicolumn{1}{r}{\cellcolor[rgb]{ .949,  .949,  .949} \textcolor[rgb]{ 0,  .439,  .753}{0.86}} & \multicolumn{1}{r}{\cellcolor[rgb]{ .949,  .949,  .949} \textcolor[rgb]{ 1,  0,  0}{\textbf{0.59}}} & \multicolumn{1}{r}{\cellcolor[rgb]{ .949,  .949,  .949} 0.69} & \multicolumn{1}{r}{\cellcolor[rgb]{ .949,  .949,  .949} \textcolor[rgb]{ 0,  .439,  .753}{0.49}} & \multicolumn{1}{r}{\cellcolor[rgb]{ .949,  .949,  .949} \textbf{0.37}} & \multicolumn{1}{r}{\cellcolor[rgb]{ .949,  .949,  .949} 0.43} & \multicolumn{1}{r}{\cellcolor[rgb]{ .949,  .949,  .949} 1.14} & \multicolumn{1}{r}{\cellcolor[rgb]{ .949,  .949,  .949} 0.77} & \multicolumn{1}{r}{\cellcolor[rgb]{ .949,  .949,  .949} 0.93} & \multicolumn{1}{r}{\cellcolor[rgb]{ .949,  .949,  .949} 0.756} \\
          &       &       & \multicolumn{1}{r}{0.43} & \multicolumn{1}{r}{\textcolor[rgb]{ 1,  0,  0}{\textbf{0.29}}} & \multicolumn{1}{r}{0.38} & \multicolumn{1}{r}{\textcolor[rgb]{ 1,  0,  0}{\textbf{0.42}}} & \multicolumn{1}{r}{0.36} & \multicolumn{1}{r}{0.34} & \multicolumn{1}{r}{\textbf{0.13}} & \multicolumn{1}{r}{0.40} & \multicolumn{1}{r}{0.42} & \multicolumn{1}{r}{0.29} & \multicolumn{1}{r}{0.57} & \multicolumn{1}{r}{0.366} \\
    \toprule
    \end{tabular}%
\vspace{-0.4cm}

\end{table*}%
 
\begin{figure}[t]
\setlength{\abovecaptionskip}{0pt} 
\setlength{\belowcaptionskip}{0pt} 
	\centering
	\includegraphics[scale = 0.65]{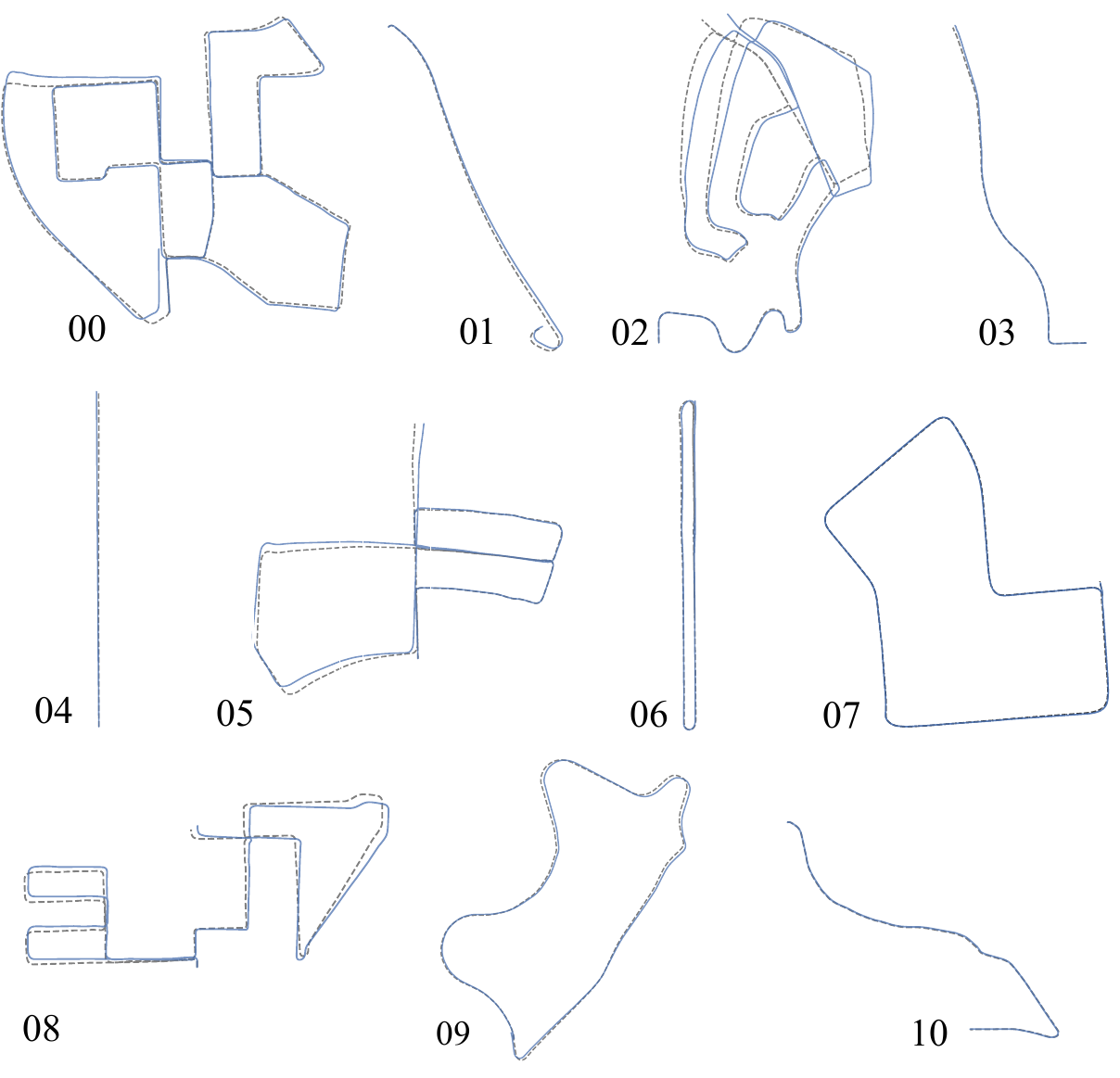}
	\caption{Trajectories estimated by our SLAMesh on the KITTI $00\sim10$ sequences (solid blue line) compared with the ground truth (dash gray line). Most trajectories are well consistent with the ground truth even without loop closures.}
	\label{figtrajectorykitti}
	\vspace{-0.4cm}
\end{figure}

We also quantitatively evaluate our pose estimation performance in the widely-used KITTI \cite{geiger2013vision} odometry benchmark. The KITTI dataset provides point clouds from a Velodyne HDL-64E LiDAR and ground-truth poses. The results on the sequences $00\sim10$ (including urban, country, and highway environments) are shown in Fig. \ref{figtrajectorykitti}. Our estimated trajectories present high consistency compared with ground truth. 

Several state-of-the-art methods with different types of maps are tested. Suma \cite{behley2018efficient} is a surfel-based method, and Suma++ \cite{chen2019suma++} enhances the odometry accuracy with dynamic object removal. LiTAMIN2 \cite{yokozuka2021litamin2} maintains a voxelized NDT map. A-LOAM is an implementation of LOAM \cite{zhang2017low}. Puma \cite{vizzo2021poisson} is also a mesh-based method. Tab. \ref{tab2kittiresult} shows the quantitative evaluation results with the standard relative translation error and rotation error \cite{geiger2013vision}. The results of these methods are directly imported from their published papers. Our SLAMesh achieves superior performance with $0.676\%$ average translation error and $0.291$deg/100m rotation error, which outperforms those point cloud, NDT, surfel, and mesh-based methods. Notice that we can achieve this without any loop closure. Also, as reported in \cite{behley2018efficient}\cite{chen2019suma++}, Suma needs GPU, and the surfel map is relatively cluttered.

Fig. \ref{figmeshsample} also displays the online mesh maps built by SLAMesh in the sequences 00, 07, and 09. The maps exhibit fine alignments when the vehicle travels back to the start. The maximum travel length of these sequences is $5.07$km, which demonstrates the scalability of SLAMesh. An interesting observation is that when we allow the current scan to register with revisited areas, the map and the whole trajectory show better consistency, but when evaluating the relative error of odometry in KITTI, disabling this implicit loop closure enables higher accuracy. We believe this phenomenon is common in methods using cell-organized maps like \cite{koide2021voxelized}.

\begin{figure}[t]
\setlength{\abovecaptionskip}{0pt} 
\setlength{\belowcaptionskip}{0pt} 
	\centering
	\includegraphics[scale = 0.50]{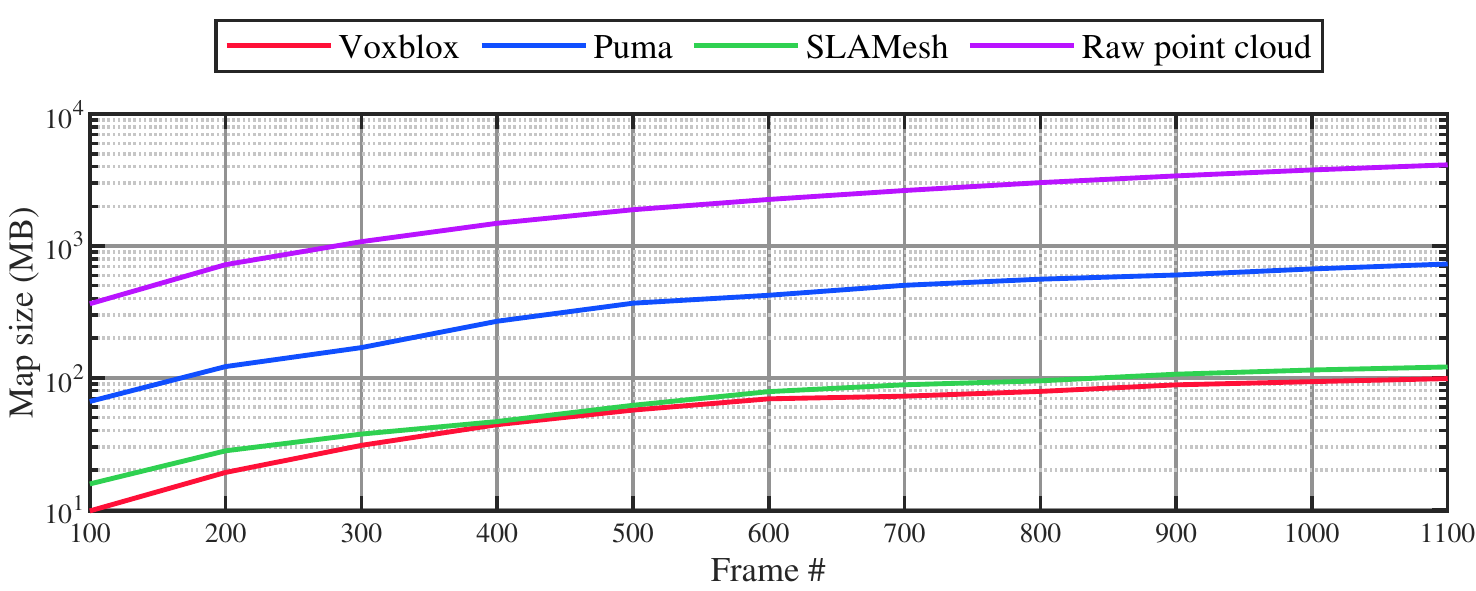}
	\caption{Memory consumption of maps with the growing number of frames on the KITTI sequence 07. The raw point cloud can occupy several GB of disk space. The three mesh-based methods have lower memory costs. However, the mesh maps produced by Puma are more complex and heavier than those produced by Voxblox and SLAMesh.}
	\label{figmemory}
 	\vspace{-0.2cm}

\end{figure}
 
\begin{figure}[t]
\setlength{\abovecaptionskip}{0pt} 
\setlength{\belowcaptionskip}{0pt} 
	\centering
	\includegraphics[scale = 0.50]{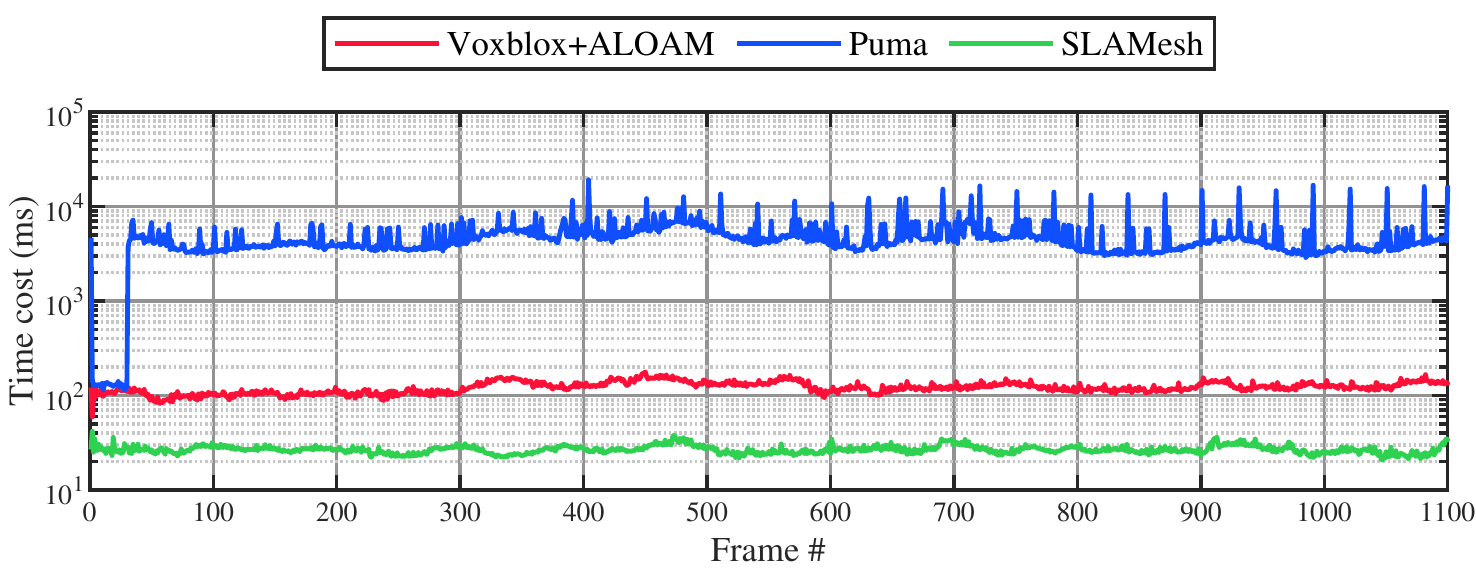}
	\caption{Time cost per frame on the KITTI sequence 07. SLAMesh can run at 40Hz. The Voxblox+A-LOAM pipeline \cite{oleynikova2017voxblox} achieves approximate real-time performance, while Puma cannot.}
	\label{figtiming}
	\vspace{-0.2cm}
\end{figure}

\subsection{Memory and Computational Efficiency}

The map size from each method in the KITTI sequence 07 is reported in Fig. \ref{figmemory}. The raw point-cloud map proliferates to several GBs, which would be too heavy for online tasks. Three mesh-based methods demonstrate their superiority in memory efficiency. Voxblox and SLAMesh consume fewer memory resources than Puma. We conjecture the reason is that Puma does not iteratively fuse point clouds.

Fig. \ref{figtiming} displays the time cost per frame on the KITTI sequence 07. The time cost of SLAMesh does not increase with the scale of the scenario due to the cell-organized map. Note that the LiDAR runs at $10$Hz in this dataset. However, Puma costs $4.7$s per frame on average. For the first 30 frames, Puma uses iterative closest point (ICP) rather than mesh. The A-LOAM+Voxblox pipeline costs $129$ms per frame ($7.8$Hz), which is approximately real-time. Our SLAMesh can run at $40$Hz and surpasses other methods. 

The time costs of each module are further analyzed in Fig. \ref{figtimebreak}. They are generally categorized into three groups. The first part, pre-process, includes down-sampling and GP reconstruction in SLAMesh, feature extraction in A-LOAM, and normal computation in Puma. The second part is registration, including data association and optimization. The last part is map maintenance, including mesh map update and publishing. In Puma, the Poisson reconstruction is the main burden. In SLAMesh and A-LOAM, registration both iterates twice. The main computational cost of SLAMesh is the pre-process (account for $63\%$), while for A-LOAM, the registration and map update cost $87\%$ time. This observation shows our different strategies for using reconstruction to structure data in advance. Voxblox is an efficient meshing tool due to its fast-integrate strategy. The max range of the sensor is cut into $50$m rather than $100$m, which also benefits its efficiency. However, the pipeline needs to maintain two maps, a point cloud map with KD-tree and a TSDF map, which costs more computation and memory resources.

\subsection{Ablation Study}

We perform an ablation study to investigate the contribution of each technique in our system. The parameter in SLAMesh is the same for all the variants. Tab. \ref{tab2kittiresult} shows the results. If SLAMesh disables the point-to-mesh error metric, the translation and rotation errors increase. This demonstrates that the pose estimation could be more accurate given more geometric information, like the normal of surfaces. The constraints combination also reduces the errors. The reason is that individual point-to-mesh error metric may not be robust enough in unstructured areas, and the constraints combination could suppress the influence of outliers. 
 
\begin{figure}[t]
\setlength{\abovecaptionskip}{0pt} 
\setlength{\belowcaptionskip}{0pt} 
	\centering
	\includegraphics[scale = 0.62]{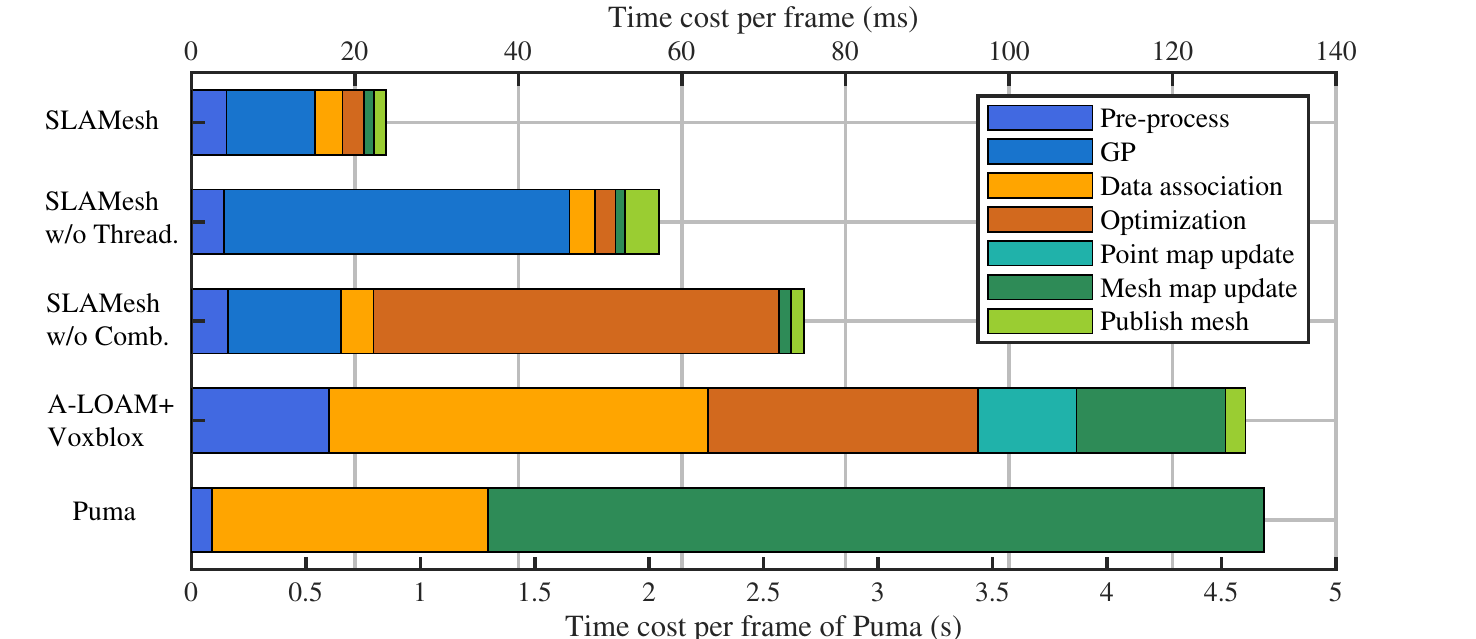}
	\caption{The time cost breakdown for each module in different methods. The main burdens in SLAMesh, Puma, and A-LOAM are GP reconstruction, mesh building, and registration, respectively. The Voxblox+A-LOAM pipeline needs to maintain two maps. The constraints combination (Comb.) and multithreading (Thread.) notably accelerate the GP-reconstruction and optimization modules in our SLAMesh. Note that Puma uses the bottom x-axis, and the other methods use the top x-axis. The figure is best viewed in color.}
	\label{figtimebreak} 
  	\vspace{-0.4cm}

\end{figure}

Constraints combination and multithreading also play essential roles in efficiency improvement. In Fig. \ref{figtimebreak}, the multithreading reduces the processing time of SLAMesh from $57.1$ms to $23.8$ms per scan. In detail, multithreading makes the GP reconstruction three times faster and the mesh publishing $2.5$ times faster. We note that Puma and Voxblox also leverage the multi-cores of the CPU in practice, which guarantees a fair comparison. The constraints combination saves about $95\%$ time cost in optimization as the average number of constraints is reduced from $9,000$ to $540$.

We conduct extensive tests with a set of parameters and find that our method has a wide tolerance for parameters. As the cell size grows to $3$m, the frame rate of SLAMesh can increase to about $80$Hz with 0.88\% relative translation error. The reason is that we use an un-parameterized model inside each voxel rather than a simple element.

 \section{Conclusions}

 We proposed a novel real-time localization and meshing system based on the reconstruction and connection strategy. Uniformly populated mesh vertices are the main property of the map, which lays a good foundation for mesh building, data association, and mesh update. Experiments showed that SLAMesh presents high odometry accuracy, efficiency, and mesh quality compared to state-of-the-art methods.

\clearpage
%
%
\bibliography{mybibfile}

\begin{thebibliography}{10}

\bibitem{zhang2017low}
Ji~Zhang and Sanjiv Singh.
\newblock Low-drift and real-time lidar odometry and mapping.
\newblock {\em Autonomous Robots}, 41(2):401--416, 2017.

\bibitem{besl1992method}
Paul~J Besl and Neil~D McKay.
\newblock Method for registration of {3-D} shapes.
\newblock In {\em Sensor fusion IV: control paradigms and data structures},
  volume 1611, pages 586--606. International Society for Optics and Photonics,
  1992.

\bibitem{vizzo2021poisson}
Ignacio Vizzo, Xieyuanli Chen, Nived Chebrolu, Jens Behley, and Cyrill
  Stachniss.
\newblock Poisson surface reconstruction for lidar odometry and mapping.
\newblock In {\em Proc. of the IEEE Intl. Conf. on Robotics \& Automation
  (ICRA)}, 2021.

\bibitem{oleynikova2017voxblox}
Helen Oleynikova, Zachary Taylor, Marius Fehr, Roland Siegwart, and Juan Nieto.
\newblock Voxblox: Incremental 3d euclidean signed distance fields for on-board
  mav planning.
\newblock In {\em 2017 IEEE/RSJ International Conference on Intelligent Robots
  and Systems (IROS)}, pages 1366--1373. IEEE, 2017.

\bibitem{hornung2013octomap}
Armin Hornung, Kai~M Wurm, Maren Bennewitz, Cyrill Stachniss, and Wolfram
  Burgard.
\newblock Octomap: An efficient probabilistic 3d mapping framework based on
  octrees.
\newblock {\em Autonomous robots}, 34(3):189--206, 2013.

\bibitem{curless1996volumetric}
Brian Curless and Marc Levoy.
\newblock A volumetric method for building complex models from range images.
\newblock In {\em Proceedings of the 23rd annual conference on Computer
  graphics and interactive techniques}, pages 303--312, 1996.

\bibitem{magnusson2007scan}
Martin Magnusson, Achim Lilienthal, and Tom Duckett.
\newblock Scan registration for autonomous mining vehicles using {3D-NDT}.
\newblock {\em Journal of Field Robotics}, 24(10):803--827, 2007.

\bibitem{yokozuka2021litamin2}
Masashi Yokozuka, Kenji Koide, Shuji Oishi, and Atsuhiko Banno.
\newblock Litamin2: Ultra light lidar-based slam using geometric approximation
  applied with kl-divergence.
\newblock In {\em 2021 IEEE International Conference on Robotics and Automation
  (ICRA)}, pages 11619--11625. IEEE, 2021.

\bibitem{behley2018efficient}
Jens Behley and Cyrill Stachniss.
\newblock Efficient surfel-based slam using 3d laser range data in urban
  environments.
\newblock In {\em Robotics: Science and Systems}, volume 2018, 2018.

\bibitem{chen2019suma++}
Xieyuanli Chen, Andres Milioto, Emanuele Palazzolo, Philippe Giguere, Jens
  Behley, and Cyrill Stachniss.
\newblock Suma++: Efficient lidar-based semantic slam.
\newblock In {\em 2019 IEEE/RSJ International Conference on Intelligent Robots
  and Systems (IROS)}, pages 4530--4537. IEEE, 2019.

\bibitem{Elasticity++}
Chanoh Park, Peyman Moghadam, Jason~L. Williams, Soohwan Kim, Sridha Sridharan,
  and Clinton Fookes.
\newblock Elasticity meets continuous-time: Map-centric dense 3d lidar slam.
\newblock {\em IEEE Transactions on Robotics}, 38(2):978--997, 2022.

\bibitem{putz2021mesh}
Sebastian P{\"u}tz, Thomas Wiemann, and Joachim Hertzberg.
\newblock The mesh tools package--introducing annotated 3d triangle maps in
  ros.
\newblock {\em Robotics and Autonomous Systems}, 138:103688, 2021.

\bibitem{chen2021range}
Xieyuanli Chen, Ignacio Vizzo, Thomas L{\"a}be, Jens Behley, and Cyrill
  Stachniss.
\newblock Range image-based lidar localization for autonomous vehicles.
\newblock {\em arXiv preprint arXiv:2105.12121}, 2021.

\bibitem{durrant2006simultaneous}
Hugh Durrant-Whyte and Tim Bailey.
\newblock Simultaneous localization and mapping: part i.
\newblock {\em IEEE robotics \& automation magazine}, 13(2):99--110, 2006.

\bibitem{rasmussen2003gaussian}
Carl~Edward Rasmussen.
\newblock Gaussian processes in machine learning.
\newblock In {\em Summer School on Machine Learning}, pages 63--71. Springer,
  2003.

\bibitem{li2020gp}
Bo~Li, Yingqiang Wang, Yu~Zhang, Wenjie Zhao, Jianyuan Ruan, and Ping Li.
\newblock {GP-SLAM}: laser-based {SLAM} approach based on regionalized
  {Gaussian} process map reconstruction.
\newblock {\em Autonomous Robots}, 44(6):947--967, 2020.

\bibitem{ruan2020gp}
Jianyuan Ruan, Bo~Li, Yinqiang Wang, and Zhou Fang.
\newblock Gp-slam+: real-time 3d lidar slam based on improved regionalized
  gaussian process map reconstruction.
\newblock In {\em 2020 IEEE/RSJ International Conference on Intelligent Robots
  and Systems (IROS)}, pages 5171--5178. IEEE, 2020.

\bibitem{lorensen1987marching}
William~E Lorensen and Harvey~E Cline.
\newblock Marching cubes: A high resolution 3d surface construction algorithm.
\newblock {\em ACM siggraph computer graphics}, 21(4):163--169, 1987.

\bibitem{newcombe2011kinectfusion}
Richard~A Newcombe, Shahram Izadi, Otmar Hilliges, David Molyneaux, David Kim,
  Andrew~J Davison, Pushmeet Kohi, Jamie Shotton, Steve Hodges, and Andrew
  Fitzgibbon.
\newblock Kinectfusion: Real-time dense surface mapping and tracking.
\newblock In {\em 2011 10th IEEE international symposium on mixed and augmented
  reality}, pages 127--136. IEEE, 2011.

\bibitem{ryde2013voxel}
Julian Ryde, Vikas Dhiman, and Robert Platt.
\newblock Voxel planes: Rapid visualization and meshification of point cloud
  ensembles.
\newblock In {\em 2013 IEEE/RSJ International Conference on Intelligent Robots
  and Systems}, pages 3731--3737. IEEE, 2013.

\bibitem{piazza2018real}
Enrico Piazza, Andrea Romanoni, and Matteo Matteucci.
\newblock Real-time cpu-based large-scale three-dimensional mesh
  reconstruction.
\newblock {\em IEEE Robotics and Automation Letters}, 3(3):1584--1591, 2018.

\bibitem{lee1980two}
Der-Tsai Lee and Bruce~J Schachter.
\newblock Two algorithms for constructing a delaunay triangulation.
\newblock {\em International Journal of Computer \& Information Sciences},
  9(3):219--242, 1980.

\bibitem{schreiberhuber2019scalablefusion}
Simon Schreiberhuber, Johann Prankl, Timothy Patten, and Markus Vincze.
\newblock Scalablefusion: High-resolution mesh-based real-time 3d
  reconstruction.
\newblock In {\em 2019 International Conference on Robotics and Automation
  (ICRA)}, pages 140--146. IEEE, 2019.

\bibitem{kuhner2020large}
Tilman K{\"u}hner and Julius K{\"u}mmerle.
\newblock Large-scale volumetric scene reconstruction using lidar.
\newblock In {\em 2020 IEEE International Conference on Robotics and Automation
  (ICRA)}, pages 6261--6267. IEEE, 2020.

\bibitem{roldao20193d}
Luis Rold{\~a}o, Raoul de~Charette, and Anne Verroust-Blondet.
\newblock 3d surface reconstruction from voxel-based lidar data.
\newblock In {\em 2019 IEEE Intelligent Transportation Systems Conference
  (ITSC)}, pages 2681--2686. IEEE, 2019.

\bibitem{kazhdan2006poisson}
Michael Kazhdan, Matthew Bolitho, and Hugues Hoppe.
\newblock Poisson surface reconstruction.
\newblock In {\em Proceedings of the fourth Eurographics symposium on Geometry
  processing}, volume~7, page~0, 2006.

\bibitem{lin2023immesh}
Jiarong Lin, Chongjiang Yuan, Yixi Cai, Haotian Li, Yuying Zou, Xiaoping Hong,
  and Fu~Zhang.
\newblock Immesh: An immediate lidar localization and meshing framework.
\newblock {\em arXiv preprint arXiv:2301.05206}, 2023.

\bibitem{Kim2015}
Soohwan Kim and Jonghyuk Kim.
\newblock {\em GPmap: A Unified Framework for Robotic Mapping Based on Sparse
  Gaussian Processes}, pages 319--332.
\newblock Springer International Publishing, Cham, 2015.

\bibitem{lee2019online}
Bhoram Lee, Clark Zhang, Zonghao Huang, and Daniel~D Lee.
\newblock Online continuous mapping using gaussian process implicit surfaces.
\newblock In {\em 2019 International Conference on Robotics and Automation
  (ICRA)}, pages 6884--6890. IEEE, 2019.

\bibitem{koide2021voxelized}
Kenji Koide, Masashi Yokozuka, Shuji Oishi, and Atsuhiko Banno.
\newblock Voxelized gicp for fast and accurate 3d point cloud registration.
\newblock In {\em 2021 IEEE International Conference on Robotics and Automation
  (ICRA)}, pages 11054--11059. IEEE, 2021.

\bibitem{knapitsch2017tanks}
Arno Knapitsch, Jaesik Park, Qian-Yi Zhou, and Vladlen Koltun.
\newblock Tanks and temples: Benchmarking large-scale scene reconstruction.
\newblock {\em ACM Transactions on Graphics (ToG)}, 36(4):1--13, 2017.

\bibitem{geiger2013vision}
Andreas Geiger, Philip Lenz, Christoph Stiller, and Raquel Urtasun.
\newblock Vision meets robotics: The kitti dataset.
\newblock {\em The International Journal of Robotics Research},
  32(11):1231--1237, 2013.

\end{thebibliography}

\end{document}